\documentclass[a4paper,fleqn]{cas-sc}

\usepackage[authoryear,longnamesfirst]{natbib}
\def\tsc#1{\csdef{#1}{\textsc{\lowercase{#1}}\xspace}}
\tsc{WGM}
\tsc{QE}

\graphicspath{{figures/}}
\begin{document}
\let\WriteBookmarks\relax
\def\floatpagepagefraction{1}
\def\textpagefraction{.001}

\shorttitle{AGI for Education}    

\shortauthors{Latif et al.}  

\title{AGI: Artificial General Intelligence for Education} 



%

\author[1,2]{Ehsan Latif}
\fnmark[1]
\ead{ehsan.latif@uga.edu}
\affiliation[1]{organization={AI4STEM Education Center},
            addressline={University of Georgia},
            city={Athens},
            state={GA},
            country={United States}}
\affiliation[2]{organization={School of Computing},
            addressline={University of Georgia},
            city={Athens},
            state={GA},
            country={United States}}

\author[5,2]{Gengchen Mai}
\fnmark[1]
\ead{gengchen.mai25@uga.edu}
\affiliation[5]{organization={Department of Geography},
            addressline={University of Georgia},
            city={Athens},
            state={GA},
            country={United States}}

\author[1]{Matthew Nyaaba}
\fnmark[1]
\ead{Matthew.Nyaaba@uga.edu}

\author[2]{Xuansheng Wu}
\ead{xuansheng.wu@uga.edu}

\author[2]{Ninghao Liu}
\ead{ninghao.liu@uga.edu}

\author[1,3]{Guoyu Lu}
\ead{guoyu.lu@uga.edu}
\affiliation[3]{organization={School of Electrical and Computer Engineering},
           addressline={University of Georgia},
            city={Athens},
            state={GA},
            country={United States}}

\author[4]{Sheng Li}
\ead{vga8uf@virginia.edu}
\affiliation[4]{organization={School of Data Science},
            addressline={University of Virginia},
            city={Charlottesville},
            state={VA},
            country={United States}}

\author[1,2]{Tianming Liu}
\ead{tliu@uga.edu}

\author[1]{Xiaoming Zhai}
\cormark[1]
\ead{xiaoming.zhai@uga.edu}

\cortext[1]{Corresponding author}



\begin{abstract}
Artificial general intelligence (AGI) has gained global recognition as a future technology due to the emergence of breakthrough large language models and chatbots such as GPT-4 and ChatGPT, respectively. AGI aims to replicate human intelligence through computer systems, one of the critical technologies that has the potential to revolutionize education. Compared to conventional AI models, typically designed for a limited range of tasks, demand significant amounts of domain-specific data for training and may not always consider intricate interpersonal dynamics in education.
AGI, driven by the recent large pre-trained models, represents a significant leap in the capability of machines to perform tasks that require human-level intelligence, such as reasoning, problem-solving, decision-making, and even understanding human emotions and social interactions. 
In this position paper, we have articulated the AGI's key concepts, capabilities, scope, and potential within future education, including achieving future educational goals, designing pedagogy and curriculum, and performing assessments.  It highlights that AGI can significantly improve intelligent tutoring systems, educational assessment, and evaluation procedures. AGI systems can adapt to individual student needs, offering tailored learning experiences. They can also provide comprehensive feedback on student performance and dynamically adjust teaching methods based on student progress. The paper emphasizes that AGI's capabilities extend to understanding human emotions and social interactions, which are critical in educational settings. The paper discusses that ethical issues in education with AGI include concerns about data bias, fairness, and privacy and emphasizes the need for codes of conduct to ensure responsible AGI use in academic settings like homework, teaching, and recruitment.
We also conclude that the development of AGI necessitates interdisciplinary collaborations between educators and AI engineers to advance research and application efforts.
\end{abstract}




\maketitle

\section{Introduction}\label{introduction}

In this era of unprecedented technological advancement, the emergence of Artificial General Intelligence (AGI) is expected to revolutionize various facets of human life, with education no exception. AGI, the ability of machines to perform tasks on par with human intelligence across a wide range of domains, has sparked a global debate about the future of education and the role of human educators. For example, the exponential growth in AGI's capabilities has raised pertinent questions regarding the nature of human learning and the necessity of adapting traditional educational models \citep{zhai2022chatgpt}. Furthermore, as AGI continues to permeate various sectors, it is essential to examine how educational systems can evolve to effectively integrate this technology, equipping students with the skills necessary to thrive in an increasingly automated world \citep{zhai2023chatgpt}. 

AI has made impressive educational strides, thanks to developments in many areas, such as machine learning (deep learning), computer vision, and natural language processing \citep{norvig2016artificial, goodfellow2016deep}. AI techniques are now widely used in various facets of daily life, such as e-commerce, healthcare, transportation, manufacturing, media, entertainment, and education. The majority of these applications in education, however, use narrow AI or specialized AI, which is created to execute particular jobs (e.g., automatic scoring, adaptive learning) with skill and frequently outperforms humans in specific restricted domains \citep{bostrom2014superintelligence, lake2017building}. Despite achieving narrow AI, creating AGI for education is still a challenging objective \citep{goertzel2014engineering}. With the help of AGI, computers can comprehend, pick up knowledge, and carry out any intellectual work that people can. It aims to develop systems with a profound grasp of the human condition and functionally similar to human intelligence \citep{voss2007essentials}. In particular, its educational implications pose several queries, difficulties, and opportunities. AGI has the potential to completely transform how teaching and learning activities are planned and carried out in the education sector. 

While the existing literature posits the transformative potential of AGI in education, a critical examination is necessary to understand its implications, challenges, and opportunities fully. Previous studies have often focused on the capabilities of AGI to personalize learning experiences and improve assessment methods. However, these claims require scrutiny in the context of real-world applications. For instance, while AGI's ability to adapt to individual learning styles is promising, it also raises concerns about data privacy and the perpetuation of existing educational inequalities \citep{holmes2021ethics}. Similarly, the touted benefits of AGI in automating assessments must be weighed against academic integrity issues, as advanced AI models could artificially produce answers for tests or assignments \citep{zhai2023ai}. 

Given the continual developments in AI and the increased interest in creating AGI systems, it is critical to investigate how AGI can change education and promote teaching and learning.
The key contributions of this position paper are as follows:
\begin{itemize}
    \item The paper aims to highlight the potential of AGI in the context of education, analyzing its implications, challenges, ethics, and opportunities that this technological breakthrough presents in the light of recent literature.
    \item The paper establishes positions in diverse research areas, including the impact of AGI on pedagogy, curriculum design, assessment methods, and the role of human educators in the face of advanced machine intelligence.
    \item By critically analyzing the existing body of knowledge, this paper aims to synthesize key findings, identify research gaps, and propose future research directions.
    \item We aim to conceptualize AGI concepts from the published studies from 2018 to 2023, although some attention is given to earlier prominent work that provides foundational knowledge of the domain. We include works mostly written in English. Most recent archived studies from renowned authors in AI and education are also considered.
\end{itemize}

This position paper will serve as a foundation for scholars, policymakers, and educators alike, fostering a robust discourse on the role of education in the AGI era and the strategies required to harness its potential while safeguarding human agency and nurturing creativity, collaboration, and critical thinking in tomorrow's learners.

\begin{figure}
    \centering
    \includegraphics[width=1\linewidth]{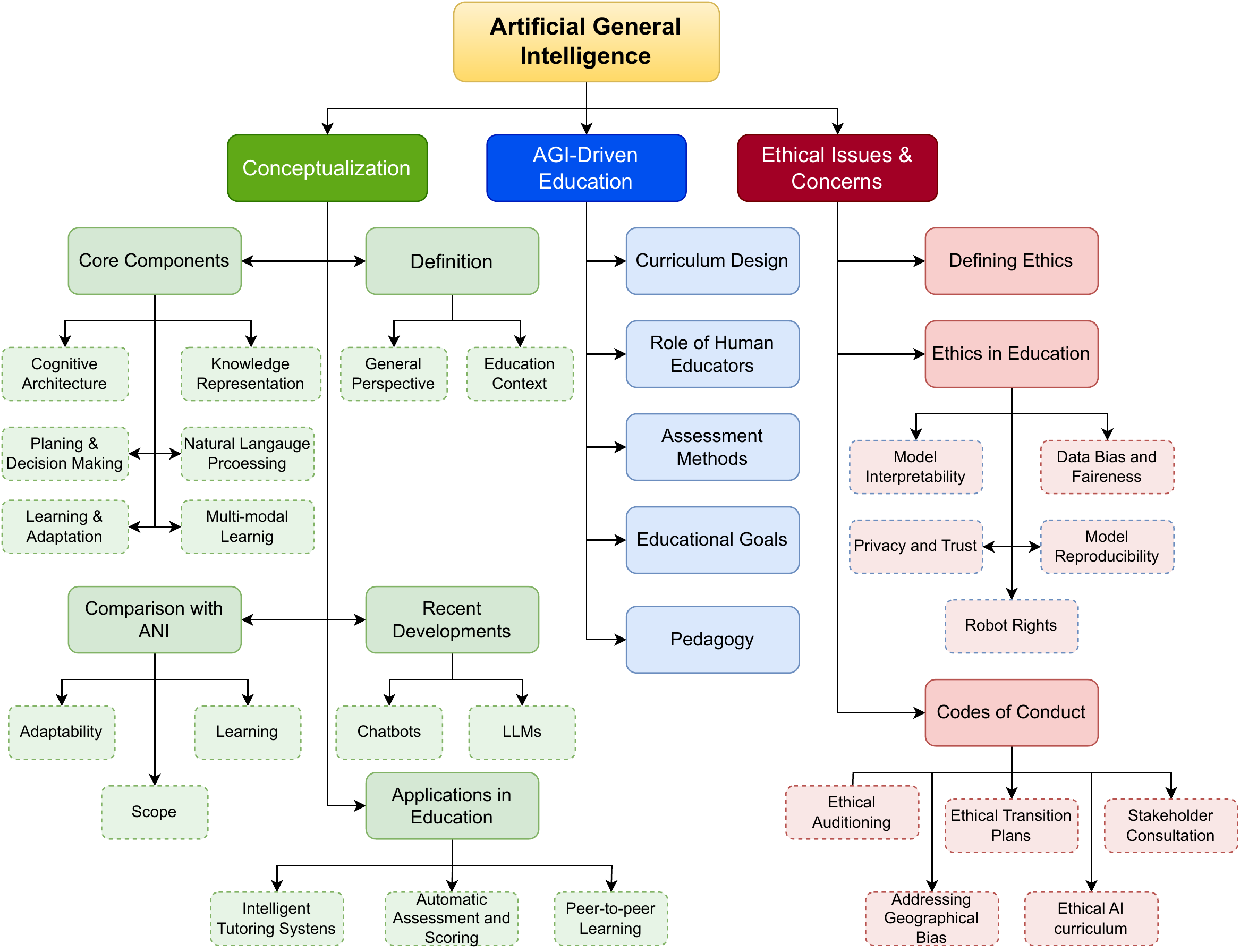}
    \caption{Overview of the AGI study in education}
    \label{fig:agi_overview}
\end{figure}

We organize the paper in an intellectual order by first conceptualizing AGI, providing a general definition, core components, comparison with conventional AI, recent developments, and educational context definition in Sec.~\ref{agi}. Sec.~\ref{potential} provides a discussion on the potential of AGI in future education, focusing on curriculum design, the role of human educators, assessment generation, and pedagogy. Lastly, in Sec.~\ref{ethics}, we provide details about ethical concerns regarding the use of AGI in education, specifically regarding data bias, model interpretation, privacy, trust, and robot rights, along with the code of conduct focuses on ethical auditing, ethical transitioning, stakeholder consultation, and Ethical curriculum design. The study overview can be seen in Fig.~\ref{fig:agi_overview}.

\section{What is AGI}\label{agi}

\subsection{Defining AGI: A General Perspective}
AGI usually refers to machine intelligence that possesses human-like cognitive abilities. For instance, an AGI agent shall be capable of understanding, learning, and carrying out any intellectual work that a human person is capable of  \citep{legg2007collection}. AGI systems mimic humans' general-purpose problem-solving abilities  \citep{wang2019defining}. The ability of AGI systems to function autonomously, making judgments and conducting actions without the need for ongoing human supervision, is one of these features. Thanks to this degree of autonomy, AGI may work well in complicated, dynamic situations, enabling it to adjust to unforeseen conditions \citep{zhai2021review}. AGI may solve problems and carry out activities in multiple domains without being restricted to a single area of competence by accumulating information and abilities in a general-purpose manner \citep{lake2017building}.

AGI is further distinguished by its ability to change and grow in response to new knowledge and evolving circumstances. AGI systems can change their behavior and can experience this adaptability, just like people do \citep{kahneman2011thinking}. Lastly, goal orientation is another characteristic of AGI systems that allows them to set and pursue goals while performing actions to achieve desired results. AGI may work strategically and plan for the future thanks to this goal-oriented behavior, which shows that it is aware of the effects and ramifications of its activities \citep{legg2007collection}. AGI research involves multiple disciplines, such as computer science, cognitive psychology, neurology, and philosophy \citep{goertzel2014engineering}. To achieve AGI, one must understand the underlying principles of human cognition and replicate these processes in computers, in addition to enhancing the processing power of AI systems. Researchers are still exploring the creation of intelligent AI systems as they work to overcome the ongoing problem of AGI. Large language models (e.g., ChatGPT and GPT-4) recently performed remarkably in language understanding, generation, and reasoning, demonstrating more general intelligence than previous AI models. Therefore, many researchers consider these large language models preliminary versions of AGI systems.  

\subsection{Core of AGI}
\begin{figure*}[t]
    \centering
\includegraphics[width=0.50\linewidth]{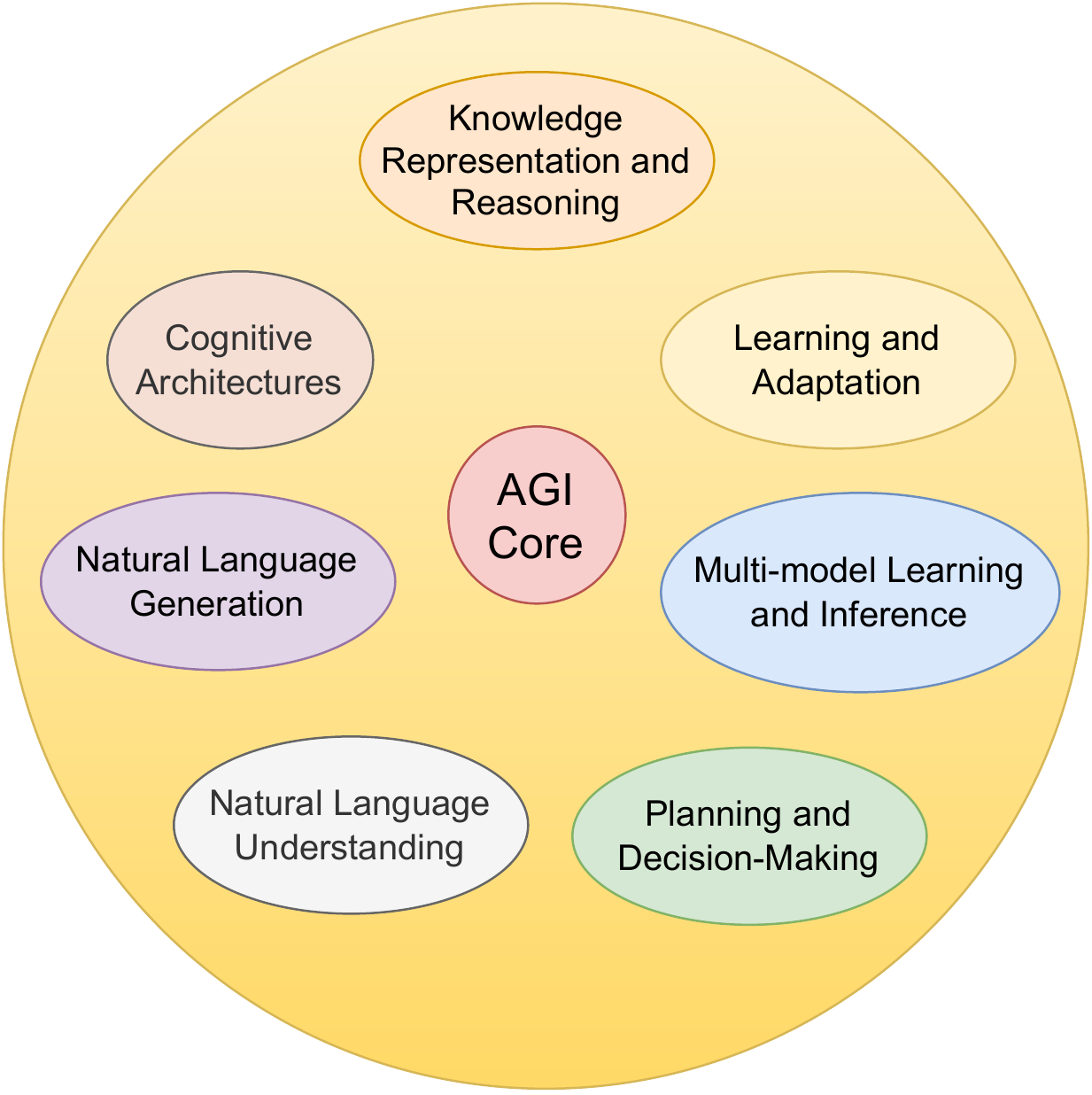}
\caption{A microscopic view of AGI Core}
    \label{fig:agi_core}
    \vspace{-2mm}
\end{figure*}

AGI is based on several fundamental ideas and processes that try to imitate general human intelligence in machines. These principles provide the foundation for creating AGI systems that can autonomously gather and process knowledge, reason, and adapt to new tasks and obstacles, which span several facets of cognition, learning, and decision-making \citep{goertzel2014engineering}. Fig.~\ref{fig:agi_core} provides a pictorial view of AGI's core with related nodes.

\textbf{Cognitive architectures:} Designing cognitive architectures, which offer a comprehensive framework for integrating various cognitive processes and modules, such as perception, memory, learning, and reasoning, is one method for creating AGI systems \citep{langley2006cognitive}. AGI systems can operate similarly to human minds thanks to these architectures, which seek to instantiate human cognition's fundamental structure and principles in computational models.

\textbf{Knowledge representation and reasoning:} One of the core components of AGI is the capacity to represent and manipulate knowledge. AGI systems must be able to store information about the outside world, use that information to reason about it to draw inferences and conclusions and update their knowledge based on new information and experiences \citep{davis2015commonsense}. Various formalisms and methods have been proposed for representing and using knowledge in AGI systems, including logic, probabilistic models, semantic networks \citep{speer2013conceptnet}, and knowledge graphs \citep{noy2019industry,janowicz2022know,qi2023evkg}.

\textbf{Learning and adaptation:} AGI systems must possess the capacity for experience-based learning and task-specific adaptation \citep{bommasani2021opportunities}. Furthermore, AGI systems can learn, modify their behavior, and improve their performance using various power tools and methods made possible by machine learning algorithms like deep learning, reinforcement learning, and unsupervised learning \citep{goodfellow2016deep}.

\textbf{Planning and decision-making:} For AGI systems to accomplish their objectives and navigate challenging circumstances, they need to be able to plan and make decisions. With tools like search, optimization, and game-theoretic methods, planning and decision-making issues can be modeled and resolved, enabling AGI systems to create and carry out plans that maximize their goals \citep{norvig2016artificial,zhai2021advancing}.

\textbf{Natural language understanding and generation:} AGI systems must comprehend and produce human languages to communicate with humans and process knowledge similar to that of humans \citep{ouyang2022instructgpt,openai2022chatgpt}. AGI systems can be given the ability to comprehend and produce text and voice using natural language processing techniques, such as syntactic, semantic, and discourse analysis \citep{zhai2022assessing}. This will improve communication and collaboration with human users \citep{hirschberg2015advances, zhai2022applying}.

\textbf{Multimodal learning and inference:}  Real-world scenarios usually involve observations in multiple modalities, such as text, image, video, audio, etc. Exploiting rich information from multimodal data is essential in building AGI systems. The AGI systems should be capable of processing multimodal inputs, extracting coherent knowledge from them, and making decisions accordingly ~\citep{zhao2023brain,lee2023multimodality}.   

These core principles and mechanisms form the foundation upon which AGI systems are built, guiding the development of intelligent agents that can operate and learn like human intelligence. 

\subsection{Artificial Narrow Intelligence vs Artificial General Intelligence}

Artificial intelligence encompasses a vast spectrum, ranging from narrow, specialized systems to broader, more generalized ones. Two primary categorizations in this spectrum are artificial narrow intelligence (ANI) and artificial intelligence (AGI).

ANI~\citep{saghiri2022survey}, often termed as Weak AI, specializes in executing a singular or a set of closely related tasks. On the other hand, AGI~\citep{mclean2023risks}, commonly known as Strong AI, alludes to machines that can comprehend, reason, and perform tasks across multiple domains at a human-like level. Its characteristics include The operational framework-based comparison as follows:
\begin{itemize}
    \item \textbf{Scope:} ANI is narrow and specialized, tailored for specific tasks like speech recognition or image classification. However, AGI is broad and versatile, capable of executing various tasks.
    \item \textbf{Learning:} ANI is confined to its domain and can evolve within its domain but cannot transfer knowledge across domains. Besides, AGI is not just domain-specific, AGI can acquire and apply knowledge across domains.
    \item \textbf{Adaptability:} ANI is limited to its predefined domain, regardless of proficiency within its domain. on the other hand, AGI is highly adaptable and can operate in new environments and tasks.
\end{itemize}

While ANI and AGI have advantages, AGI stands out as the clear winner, particularly in educational settings. AGI is a powerful tool for individualized learning, responding to the needs of each unique learner and providing comprehensive educational experiences because of its capacity to adapt, learn across domains, and manage various activities concurrently. However, the specialized nature of ANI might limit its application to certain instructional resources or tasks. With the ability to personalize content and methods to specific learner profiles, AGI can transform educational settings and outperform the more restricted uses of ANI. 

Generally, an AI system's classification as ANI or AGI depends on its adaptability, breadth of knowledge, and learning mechanism—even if it is initially created for a particular course or task. Its intended use may provide some initial clues, but a more definitive differentiation can be made by paying attention to its behavior, adaptability, and depth versus breadth of function.

\subsection{AGI in the Modern Era: The Development of Large Language Models}
AGI's general-purpose character and capacity for adaptation to various tasks and obstacles underpin its broad range of applications and areas. AGI systems may learn, reason, and make judgments across different domains, which enables them to be deployed in various situations and sectors. AGI systems are not constrained to specific fields or skill areas. The ChatGPT \citep{openai2022chatgpt}, built on the GPT architecture, is one well-known example of a system gradually moving towards AGI \citep{brown2020language}. ChatGPT exhibits unique abilities in natural language comprehension, inference, and generation. It demonstrates verbal proficiency suggesting a step towards AGI by engaging in context-aware, human-like conversations and producing clear, educational, and pertinent responses. The versatility of large language models (LLMs) like GPT-4 in the natural language domain makes them particularly interesting \citep{brown2020language}. These jobs include, among others, text summarization, machine translation, problem-solving, sentiment analysis, question-answering, and creative writing. As they can swiftly adjust to new tasks and obstacles in their domains without needing task-specific training, LLMs exhibit adaptability more akin to AGI systems than narrow AI \citep{wei2021finetuned,latif2024g}.

Creating LLMs (e.g., GPT-4) has educational consequences because these models can support instructors, students, and administrators in various educational scenarios. LLMs, for instance, can be used to create intelligent tutoring systems, offer feedback on student performance, produce educational content, and promote peer-to-peer learning and cooperation. One may pick a few examples from the work of \cite{mohammed2019towards} and \cite{zhai2023chatgpt1}. As researchers work to enhance AI systems' general-purpose learning and reasoning capabilities, the application of AGI in education and other fields is steadily growing.

\subsection{Defining AGI in the Context of Education}
AGI is the use of computer intelligence that resembles human intellect in education to enhance learning and educational experiences. AGI's abilities, such as its aptitude for \textit{general-purpose problem-solving}, enable it to take on various educational tasks without having received specialized training for each. Furthermore, it can mimic human-educator thought processes, including reasoning and comprehension, thanks to its \textit{human-like cognitive capacities} for automatic assessments and recommendations.

\begin{figure*}[t]
    \centering
\includegraphics[width=0.99\linewidth]{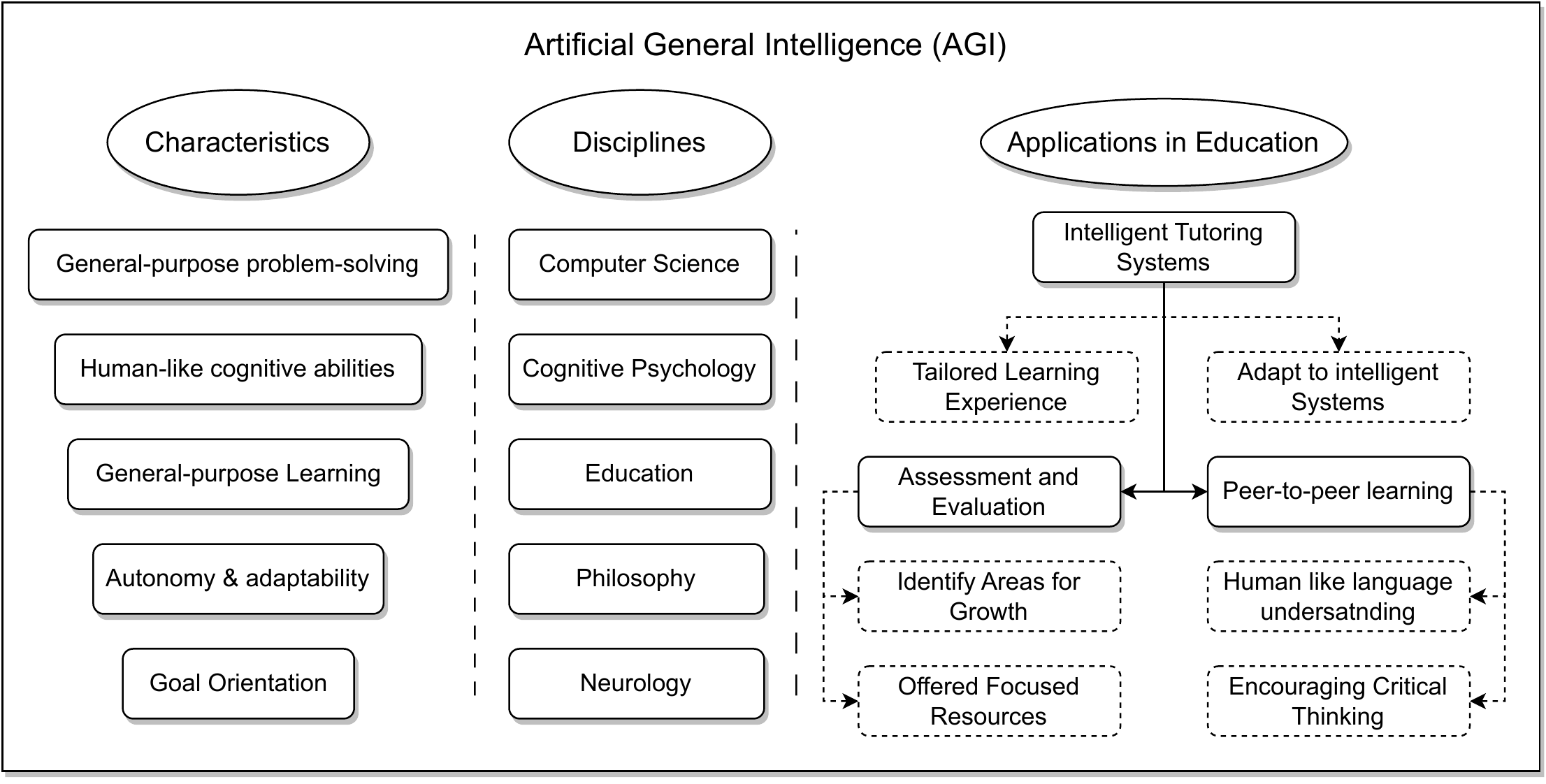}
\caption{A High-level Perspective: Artificial General Intelligence; Characteristics, Disciplines and applications}
    \label{fig:agi}
    \vspace{-2mm}
\end{figure*}

In the educational realm, AGI finds application in \textit{Intelligent Tutoring Systems}, which are personalized platforms adapting resources based on individual learners' requirements. The \textit{Tailored Learning Experience} ensures that AGI offers tasks and resources suited to individual preferences. These systems can evolve based on interactions, showcasing their adaptability. The promotion of \textit{peer-to-peer learning} fosters collaborative endeavors, allowing learners to gain knowledge from each other. Advanced student evaluations can be conducted by AGI, providing in-depth insights into learning trajectories and areas needing improvement. With its ability for \textit{human-like language understanding}, AGI can comprehend and respond in natural language, ensuring smooth interactions. Lastly, by posing challenging scenarios, AGI plays a pivotal role in \textit{encouraging critical thinking} and honing problem-solving skills among learners. Fig.\ref{fig:agi} provided an overall perspective of AGI's characteristics, disciplines, and applications in education. 

AGI has the potential to fundamentally alter how teaching and learning processes are created, put into practice, and assessed in education. AGI-driven educational systems can take advantage of AGI's wide range of cognitive capabilities and adaptability to better comprehend individual students, meet their specific needs, and create tailored learning experiences \citep{mohammed2019towards}. The autonomy and goal-oriented behavior of AGI can be used by ITS to dynamically modify the pace, material, and teaching tactics according to the student's progress, preferences, and learning styles, fostering a more enjoyable and productive learning environment \citep{graesser2005autotutor}.

For \textit{Intelligent Tutoring Systems}, AGI can improve educational assessment and evaluation procedures. AGI-driven systems can comprehend and evaluate complex material, identify areas for growth, and offer focused resources to assist students in strengthening their deficiencies. They may provide comprehensive and meaningful feedback on student performance \citep{shute2011stealth}. AGI can also learn from its interactions with students, which helps it develop its teaching abilities over time. In addition, AGI would help define new ways of evaluating student performance beyond scoring and help produce a personalized assessment framework.    By staying current with the most recent pedagogical research, educational trends, and best practices, AGI-based educational technologies may ensure that students receive high-caliber training by the most recent standards \citep{woolf2010building}.

Last but not least, AGI's capacity for understanding and producing human-like language can aid in creating systems that promote \textit{Peer-to-Peer Learning} and cooperation, imitating real-life human interactions and encouraging learners to use critical thinking and problem-solving abilities \citep{rosenschein1994rules}. In conclusion, using AGI in education can result in individualized, flexible, and successful learning experiences that better prepare students for a constantly changing global environment.

\section{The Potentials of AGI in Transforming Future Education}\label{potential}
The recent development of AGI tools' such as ChatGPT and GPT-4 have increased awareness about using digital resources in K–12 and higher education to prepare the next generation of students to solve problems of the 21st century \citep{woolf2015ai,eysenbach2023role}. The use of AGI, specifically at the K-12 level, is still in its budding stage and lacks substantial practical and empirical evidence for classroom implementation. However, a growing body of articles and research addresses the potential impact of AGI tools on enhancing teaching and learning while educators and researchers continue to explore the evidenced-based implementations of AGI. 

As AGI technologies continue to develop and mature, more research is expected to be conducted on their practical applications in the classroom \citep{baidoo2023education}. This section synthesized ideas from various articles and research papers as crucial in understanding the full potential of AGI in transforming education and preparing students for a rapidly changing world. The key concepts identified and discussed on AGI and education include its contribution to achieving educational goals, enhancing teachers' content knowledge, facilitating individualized and adaptive learning, differentiating instruction and assessment, improving student educational outcomes, etc. 

\subsection{Educational Goals}
Ensuring the high achievement of educational goals is crucial in any learning environment. Education aims to accomplish the intended learning objectives and outcomes for students. However, these goals may not be achieved due to various challenges, such as inadequate resources, limited access to skilled educators, or a lack of individualized attention for students with diverse learning needs. Nonetheless, \cite{pedro2019artificial} posited that AGI can revolutionize learning and improve learning outcomes. Their study analyzes how AGI can improve learning outcomes, particularly in developing countries. Their finding showed that AGI technology could help education systems by using data to improve educational equity and quality. For instance, AGI can analyze student performance, attendance, and engagement data to identify areas where intervention is needed, such as providing additional resources for struggling students and factoring these in setting realistic goals \citep{emerson2023multimodal}. This can help to ensure that students receive the support they need to succeed, regardless of their background or socioeconomic status.

This is to say, with available data (such as report cards, cumulative records, chief examiners' reports, etc), AGI can analyze students’ historical and current performance, creating a profile that highlights their strengths, weaknesses, learning pace, and preferences \citep{george2023managing,latif2024fine}. With this knowledge, AGI can help educators set educational goals that stretch students’ abilities and are feasible and attainable \citep{bozkurt2023speculative}. Also, AGI can assist educators in evaluating student resources, be it time, study materials, or external help like resource persons, and suggest goals that can realistically be achieved with those resources.

Ultimately, integrating potential AGI tools like chatbots in education can enhance the overall quality of education and contribute to achieving the United Nations' Sustainable Development Goals \citep{zhai2023chatgpt,farrokhnia2023swot}.
Also, they have the potential to make a significant contribution towards achieving Sustainable Development Goal 4 (SDG4), which is the education-related goal of the United Nations 2030 Agenda for Sustainable Development. Adopted in September 2015, SDGs Goal 4 aims to ensure inclusive and equitable quality education for all, promoting lifelong learning opportunities that can benefit individuals and communities alike \citep{kopnina2020education}. Potential AGI-like chatbots can help to achieve this goal by offering personalized and accessible learning experiences that cater to the needs of learners from diverse backgrounds \citep{zhai2023chatgpt1}. Potential AGI-like chatbots can provide a more engaging and interactive learning environment, helping to improve educational outcomes and support lifelong learning. 

\subsection{Pedagogy}
AGI may improve teachers' pedagogical content knowledge to promote differentiated instruction (a critical aspect of contemporary education). One of the challenges teachers face is recognizing and accommodating individual students’ needs in the classroom. AGI has the potential to greatly assist teachers in this area by providing them with a variety of subject matter knowledge, individualized lesson plans, instructional activities, and resources that can be customized to each student's unique needs.  Teachers' content knowledge encompasses the mastery of theories, facts, concepts, and skills in a specific subject area. Enhancing teachers' content knowledge is crucial for ensuring quality teaching. For instance, \cite{yang2020practical} study exploring ways to improve teaching effectiveness through AI technology' revealed that AI-based strategies have good references that significantly improve the content knowledge of courses and keep teachers up-to-date. This potential of AGI has a greater tendency to improve teachers' confidence in teaching. Students may also benefit from the wealth of knowledge \citep{zhai2023can} presented by teachers through the proper understanding of subject areas from the new insights embedded into their teaching practices. 

Nevertheless, some authors believe that AGI chatbots may not provide sufficient resources to address all learning needs \citep{gupta2023chatgpt}. Despite this criticism, the potential of AGI to promote differentiated instruction is a viable area for more exploration because teachers can access a wealth of information and activities that cater to various learning styles and preferences. This may enable teachers to develop more targeted lesson plans and strategies, ensuring each student receives the appropriate support and guidance.

It is, however, recommended that educators ensure academic integrity by preventing cheating and plagiarism. This may involve using plagiarism detection software to monitor student submissions and establishing explicit guidelines for appropriately using the model in their tests and assignments, particularly \citep{atlas2023chatgpt}. In addition to this concern, educators should emphasize the importance of proper attribution when students rely on AGI for information or inspiration, ensuring they understand how to cite the tool correctly in their work. If this is fostered, teachers can use AGI well while upholding academic standards.

\subsection{Curriculum design}
AGI can be a valuable hub for various learning resources, making it a valuable tool for designing curricula. One of the significant challenges in education is the lack of adequate resources to meet the diverse needs of students in the classroom. While there is no evidence of an AI-specific curriculum for K-12 education, \cite{chiu2020sustainable} provides valuable insights into developing a formal AI curriculum for K-12 schools. The study shows that a genuine curriculum design should encompass four forms of the design approach: discipline-specific design, learner-centered design, career-oriented design, and society-oriented design. These forms of design approach may help to ensure that the AI curriculum is comprehensive and caters to the diverse needs of K-12 students. The proposed AI-specific curriculum development is crucial as it will provide teachers and curriculum officers with the necessary tools to prepare, implement, and refine an AI curriculum that meets the four forms of the design approach. Fundamentally, \cite{chiu2020sustainable} emphasizes the importance of a specialized AI curriculum that considers the unique needs of K-12 students and provides teachers with the necessary tools to deliver quality education. 

\cite{gupta2023chatgpt} conducts a typical example of teachable curriculum design by AGI, where AGI was used to design course outlines that were engaging and interactive. The study suggests that the generated output produced by AGI contains all the necessary information for creating a course outline for the topic "Knowledge and Curriculum". The components, such as course overview, course objectives, activities, assessments, etc.,  were generated by AGI. However, the authors caution that while AGI has the potential to be a valuable tool for educators, it should be used with proper monitoring and prevention measures in place. Without proper oversight, using AI-generated course outlines may have unintended consequences and negatively impact the educational process. Therefore, educators must approach this technology cautiously and implement appropriate safeguards to ensure its effective and responsible use in the classroom.

Additionally, AGI can provide specific learning resources that can be adapted or adopted for classroom use by teachers. AGI can provide teachers with valuable resources to assist them with lesson planning, especially regarding content knowledge and assessment strategies \citep{cooper2023examining}. With AGI, Teachers will have access to a variety of content to enrich themselves and assist their students in broadening their perspectives. AGI can be considered as a supplement to required textbooks and resources that will liberate teachers from textbook limitations and widen their content knowledge \citep{cooper2023examining}. For instance,\cite{cooper2023examining} designed a challenging learner-centered lesson for seventh-grade science students using AGI. The assessment of the lesson proved that students grasped a solid understanding of renewable and nonrenewable energy sources \citep{cooper2023examining}. The fascinating aspect is that AGI  can generate assessment tasks, scoring rubrics, and a planned lesson \citep{zhai2023chatgpt1}.

AGI can, in some sense, be considered a teacher's guide or assistant, as it can potentially support teachers in various aspects of their work, such as lesson planning and assessment \citep{zhai2023chatgpt}. AGI can offer a step-by-step procedure for preparing and delivering lessons to a specific group of students based on their individual needs, preferences, and learning styles. By leveraging the power of AGI, educators can create more personalized and compelling learning experiences that cater to each student's needs, ultimately enhancing the overall teaching and learning process \citep{zhai2023chatgpt,lee2023gemini}.

Interestingly,  AGI has already started transforming how libraries provide reference services to their patrons. With their 24/7 availability, AGI can assist patrons at any time of the day or night and as also useful for those who work late hours or live in different time zones \citep{chen2023chatgpt}. In addition, AGI can be integrated with library systems, such as the library catalog, to help patrons find the information they need quickly and easily. This can lead to a more efficient and effective search process, saving patrons valuable time and effort.
Furthermore, \cite{chen2023chatgpt} indicated that potential AGI could handle multiple languages, making it an ideal solution for libraries with diverse patron populations. Language barriers can often be a significant obstacle for patrons seeking assistance, but with AGI, language is no longer limited. \cite{chen2023chatgpt} believe that AGI will become even more sophisticated and versatile with time, providing patrons a more personalized and convenient experience.

\subsection{Assessment methods}
Assessment tools are widely applied for diverse educational purposes \citep{newton2007clarifying}. 
One major purpose of assessments during teaching is guidance, which makes the students' thinking, knowledge, and abilities visible to both teachers and themselves so that the teachers can select appropriate instructional strategies to help students get closer to the educational goals in the future \citep{national2001knowing}. 
Classical assessment tools include homework, quizzes, exams, essays, projects, etc. 
Traditionally, conducting assessments was time-consuming as instructors had to be deeply involved in constructing response items and rubrics and grading student responses \citep{frey2013modern}. 
However, with the advancements of AGI technology, this process will become much more efficient. 
The following will discuss how AGI can help human teachers design and deploy assessments. 

\emph{AGI assisting teachers in designing assessment tools.}
Creating an assessment tool begins with establishing a milestone at the current stage of educational progress, followed by identifying key points to achieve this milestone. 
Teachers select milestones and critical points based on the materials used. 
For example, at the end of the first month of a semester, the milestone may be the third chapter of the textbook, and the teacher may select keywords from the first three chapters. 
While milestone selection may still rely heavily on teacher experiences, AGI systems can automatically highlight relevant content within the first three chapters as potential key points. 
The teacher can then review and refine these key points, removing any deemed unimportant to students. 
AGI systems can also assist with the design of response items and the generation of rubrics \citep{gardner2021artificial}. 
This capability could be further enhanced by the multimodal generalizability of AGI models, e.g., generating assessment items with visual and textual information \citep{fei2022towards}.
Teachers no longer need to design items themselves or search the internet; instead, they can revise the items generated by AGI systems. 
Teachers can save significant time and energy by supervising the AGI systems' generation process. 

\emph{AGI assisting teachers in deploying assessment tools.}
There are two steps to deploying an assessment tool: collecting and evaluating student responses.
Typically, the collection process involves printing the assessment items on the papers and launching a quiz in the classroom. 
However, teachers now have the option to ask the AGI systems to create a small software for the quiz by automatically self-coding, breaking free from physical constraints, and making the delivery and collecting process easier and more accessible to students. 
On the other hand, while auto-scoring systems \citep{haudek2012they,riordan2020empirical,wu2023matching} have been studied for decades to help teachers in evaluating responses, they can only produce \emph{scores} and not provide informative feedback to teachers and students. 
This single and straightforward feedback format limits the effectiveness of utilizing the results of the assessment tools.
Fortunately, new AGI systems can overcome this limitation with their strong generation ability. 
In addition to providing scores, they can provide personalized guidance to students by explaining the reasons for those scores in text format. 
Teachers can use AGI systems to summarize the explanations and AGIn an overall understanding of the class. 

Overall, AGI systems' comprehensive general knowledge and strong generative ability could let them become a human-like teaching assistant in the traditional assessment process. 
This change provides more informative feedback to both the teachers and students in a more efficient way and further enhances the impacts of assessment on improving teaching quality. 

\begin{figure*}[t]
    \centering
\includegraphics[width=0.80\linewidth]{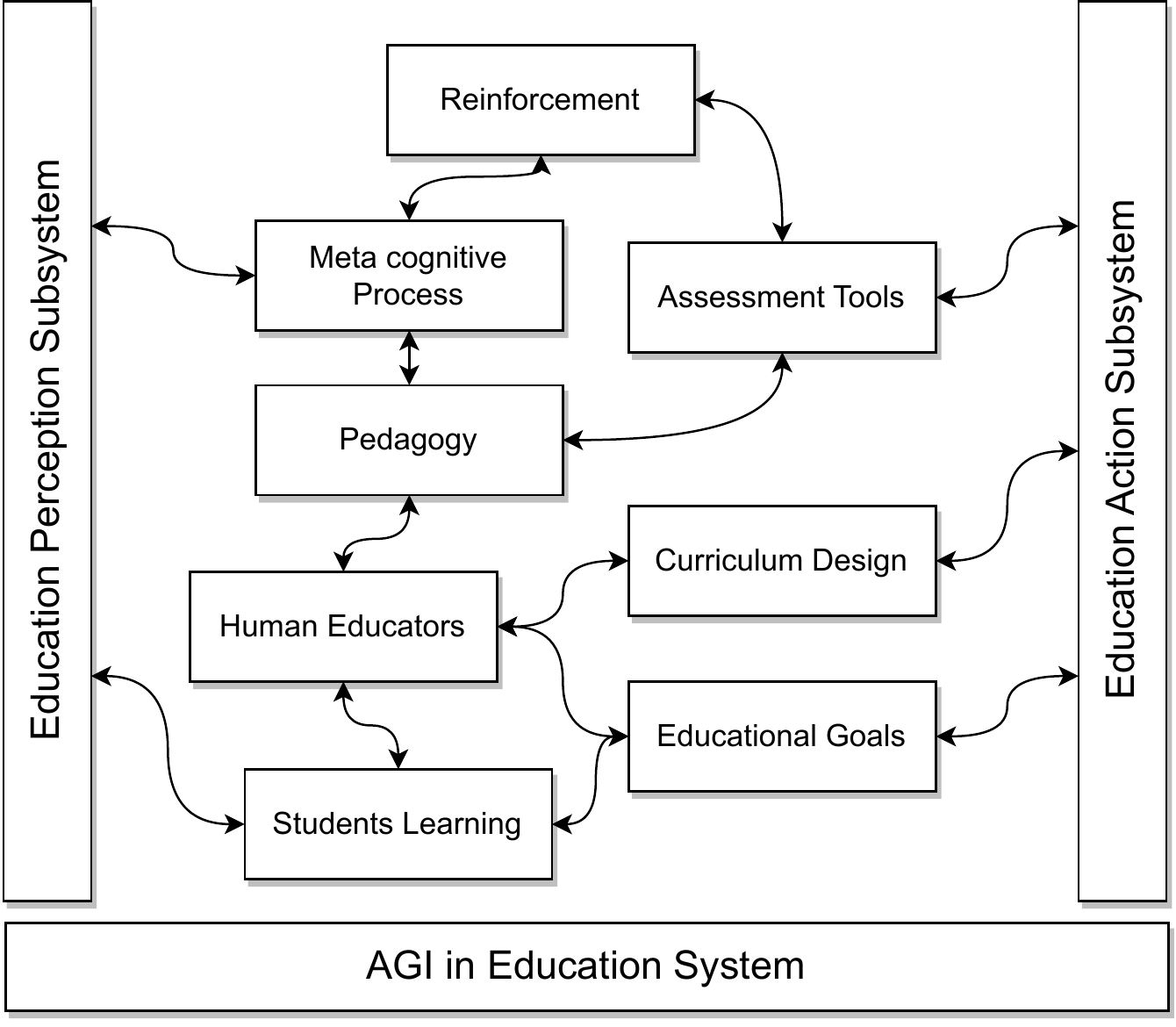}
\caption{A High-level Structure of AGI role in education with sub-systems, components, and their mapping }
    \label{fig:agi_education}
    \vspace{-2mm}
\end{figure*}

\subsection{Role of human educators}
It is essential to consider the value of human interaction and expertise in the educational process \citep{zhai2023chatgpt1}; Some concerns have been raised that AGI technology like ChatGPT and GPT-4 could potentially replace teachers or contribute to teacher shortages; . AGI technology can undoubtedly offer invaluable support and resources for educators. Therefore, it is believed that the development of independent AGI has the potential to substitute teachers soon. If it were to come to fruition, this possibility would cause significant disruption in the educational landscape, including the loss of many jobs related to teacher roles \citep{edwards2018not}. Other authors (e.g., \citep{atlas2023chatgpt}) argue that the role of a teacher extends far beyond simply imparting knowledge. Teachers provide their student's guidance, motivation, and emotional support, aspects that AGI cannot fully replicate. Therefore, AGI should be viewed as a tool that complements and enhances a teacher's capabilities rather than a threat to their profession.

Also, AGI can perform the role of content experts as it can offer a wealth of knowledge to teachers and even students. However, in education, it is highly recommended that the content provided be developmentally appropriate for the learners, considering the various psychological development levels of students. Therefore, since AGI tools, human educators are expected to fine-tune and ask specific topics-related questions to help guide AGI tools to provide responses that better align with learners' developmental stage \citep{zhai2023chatgpt,lee2024applying}. The most interesting part is that AGI is not limited to a few subject areas but is versatile in addressing various subjects, including arithmetic, science, social studies, and language arts, making it an invaluable resource for all educational levels.

Furthermore, AGI can perform the role of an assistant teacher by helping educators to provide personalized feedback, guidance, and support to students. AGI systems can also adapt to each student's pace and progress, providing remedial or advanced content as needed. For instance, AGI systems can provide cognitive clues to students, such as gaze \citep{parikh2019looking,parikh2022multi,lu2021unsupervised} and expression \citep{lu2021multi}. Teachers can utilize these clues to obtain instant insights into the cognitive states of their students, which can provide valuable information about their learning progress, and modify their teaching pace or materials, offer students personalized feedback, and implement prompt measures to improve their teaching efficacy. 

Students can also utilize AI to enrich their personalized learning experiences by analyzing their learning patterns, preferences, and needs and providing customized content and activities. This can improve engagement, retention, and academic performance. Learners require consistent communication and activities to maintain engagement, connection, and a sense of belonging to the curriculum. In such scenarios, Chatbots can be highly beneficial by offering regular updates and suggestions related to the course material. By tailoring the information and guidance to the specific needs of each student, Chatbots can ensure that learners stay informed and invested in the learning process while also monitoring their advancement. Fig.~\ref{fig:agi_education} refers to mapping the role of AGI in education with subsystems and components to achieve education goals through perception and action.

\section{Ethical Issues and Concerns}\label{ethics}
\subsection{What is Ethics?}
Ethics are a set of moral principles or codes of conduct that guide and govern human behavior in daily life \citep{proctor1998ethics}.
Ethics is the moral compass of human behavior, as old as philosophy \citep{janowicz2023philosophical}. Here, we quote \cite{janowicz2023philosophical} on \cite{jonas1984imperative} \textit{ethic of responsibility} in which Jonas reformulates Kant’s initial categorical imperative by stating that ``[we should act] so that the effects of [our] action[s] are compatible with the permanence of genuine human life''. In the science domain, as \cite{nelson2022accelerating} said, ethics typically includes reflection upon moral questions that arise in research, publication, data collection, data analysis, model development, model evaluation, and other professional and scientific activities. 

As pointed out by Janowicz \cite{janowicz2023philosophical}, while most of the work on research ethics is generated from fields such as biology, medicine, cognitive science, and social science, almost all domains of study benefit from the study of ethics. With the development of new technology and methods, many new branches of domain-specific ethics have arisen. Some good examples are BioEthics \citep{gazzaniga2005ethical}, GeoEthics \citep{goodchild2021geoethics}, and AI Ethics \citep{jobin2019aiethics}.

\subsection{AI/AGI Ethics and Their Implications in Education}

Despite all these successful stories of AI, as its applications spread to every corner of our lives, many ethical issues have arisen. 
The ethics of AI in education is a rapidly growing field, and the prior research has tackled many complex issues. One of the seminal works in this area is by \cite{holmes2021ethics}, which serves as a cornerstone in understanding the ethical dimensions of AI in educational settings, emphasizing the need for a well-designed ethical framework that addresses fairness, accountability, transparency, bias, and inclusion. It also presents the survey results among leading AI in Education (AIED) researchers, highlighting the community's need for ethical guidelines. On the other hand, \cite{zhang2023integrating, williams2023ai+} focus on the pedagogical aspect. These works explore the design and implementation of AI literacy programs for middle school students, integrating ethical considerations into the curriculum and demonstrating the effectiveness of such an integrated approach in fostering technical and ethical AI literacy.

The field is rich in scope but still in its nascent stages, calling for more comprehensive research to develop robust guidelines and best practices, especially toward AGI. For example, responsible machine learning principles are advocated as a practical framework to guide AI technologists in developing machine learning models responsibly. Eight principles have been proposed aiming at minimizing the risks of AI technology and enlarging the benefits of AI to the public by ensuring the ethical and conscious development of AI models \citep{ethicalInstituteEthical}: human augmentation, bias evaluation, explainability by justification, reproducible operations, displacement strategy, practical accuracy, trust by privacy, and data risk awareness.
In the following, we will discuss these principles.

\subsubsection{Data Bias and Fairness}

There are also numerous examples in the biomedical field where data bias has caused a lot of trouble. One exciting example by \cite{nabi2018bioethics} pointed out is a machine learning-based decision support system for predicting the probability of death in 14,199 patients with pneumonia. The model mistakenly classified patients with concomitant pneumonia and asthma as low risk while those only with pneumonia as high risk, even though the former should be more serious. The reason lies in the data collection process - patients with both diseases were admitted directly to the intensive care unit (ICU), leading to better overall outcomes. Junaid Nabi treated this as an example to show that a lack of clinical context in the AI model design results in a technically correct conclusion but based on misinterpretations of the clinical scenario. However, we want to point it out that this is essentially a data bias issue in which the patient records with both diseases are only collected from ICU cases but ignoring the non-ICU cases. This rings the bell for the awareness of data bias in machine learning models' training data collection process. 

When we apply AI models to the geospatial domain, we need to consider 
Geographic bias.  
At many times, although unlabeled geospatial data such as satellite images have wide geographic coverage and a balanced geographic distribution, the labeled geospatial data are highly geographically biased. For example, as one of the largest building footprint segmentation datasets, SpaceNet 6 \citep{shermeyer2020spacenet6} only covers 120km$^2$ over the port of Rotterdam, the Netherlands. 
The iNaturelist 2018 dataset \citep{mac2019presence}, as one of the largest geotagged images of various species, has much more images in the USA than in South America and Africa.
In this case, developing a self-supervised or unsupervised training objective on the unlabeled geospatial data \citep{mai2023csp} can mitigate the geographic bias issue inherited by AI models.
At other times, even unlabeled geospatial data are highly geographically biased. 
\cite{janowicz2016moon} visualized the spatial footprints of 15 million geographic features extracted from multiple popular Linked Data sources such as DBpedia \citep{auer2007dbpedia}, Geonames \citep{ahlers2013assessment}, Freebase, etc. The resulting map shows that even after combining multiple data sources, the integrated geographic feature sets still demonstrate a very clear geographic bias -- there are much more data in North America and Europe. In contrast, fewer data points (i.e., geographic features) are available in Africa and South America. Later on, \cite{liu2022geoparsing} show that after training on the geographically biased datasets, the commonly used machine learning-based geoparsers such as Edinburgh geoparser \citep{grover2010use} and CamCoder \citep{gritta2018melbourne} are also highly geographically biased -- they perform very well on certain regions but remain poor in other regions. How to mitigate geographic bias when training AI models is an active research area. 

While bias is harmful, \cite{zhai2022pseudo} issues a cautionary note to the field, urging researchers and practitioners to be vigilant about Pseudo Artificial Intelligence Bias (PAIB). They argue that this form of bias, widely disseminated in academic literature and public discourse, can have detrimental social impacts. Specifically, they identify three types of PAIB: misunderstandings about AI functionalities, pseudo-mechanical biases that falsely attribute inherent limitations to AI, and over-expectations that inflate what AI can realistically achieve. Zhai and Krajcik point out that these biases fuel unnecessary fears about AI perpetuate existing social inequities, and squander valuable resources invested in AI research. To counter these issues, they propose a multi-pronged approach. This includes the certification of users for AI applications to alleviate AI-related fears, offering customized user guidance to ensure ethical and effective AI usage, and the development of systematic methods to monitor and rectify bias in AI systems. The authors conclude by emphasizing the social harm caused by PAIB and call for concerted efforts to mitigate its effects for the responsible advancement of AI. Further, \cite{latif2023ai} provided evidence to support the idea of PAIB on the basis AI disparity, bias and fairness for automatic scoring.

\subsubsection{Privacy and Trustiness}
As AI technology becomes increasingly integrated into educational systems, concerns about privacy and trustworthiness have come to the forefront. For instance, \cite{mcstay2020emotional} delves into the privacy implications of emotional AI practices that do not identify individuals. They highlight that even non-identifying emotional AI can have privacy implications, emphasizing the need for a societal agreement on the principles of practice regarding privacy and the use of data about emotions. This underscores the importance of protecting the privacy of students and educators when implementing AI technologies in educational settings.

In summary, as AI and GAI technologies become more pervasive in educational settings, it is imperative to address privacy and trustworthiness proactively. Doing so protects the rights and well-being of students and educators and fosters a more responsible and ethical use of AI in education. Therefore, protecting privacy and improving trustworthiness should be integral to any AI implementation strategy in educational settings.

\subsubsection{Model Interpretability}
The need for explainability and interpretability becomes paramount as AI systems become more complex and influential in educational settings. Explainable Artificial Intelligence (XAI) \citep{gunning2019xai} is a rising ML research field aiming to explain machine decisions and predictions and justify their reliability. According to \cite{conati2018ai}, interpretability in AI is crucial for Open Learner Modelling, a branch of Intelligent Tutoring Systems. The authors argue that making AI models interpretable is essential for maximizing their educational impact and ensuring ethical considerations are met. This aligns with the broader call for explainable AI across various high-stakes domains, as noted in the work by \cite{gade2019explainable}, which emphasizes the importance of model transparency in areas like healthcare and education.

In summary, the literature strongly supports the need for explainability in AI applications in education. When ChatGPT generates an answer, we would like to know which data source its prediction is based on, so-called question answering grounding \citep{carta2023grounding}. Without this grounding process, the fidelity or truthfulness of the generated answers can not be guaranteed. Many large language models are criticized for generating inaccurate results \citep{ouyang2022instructgpt}. So from an educational perspective, it is very important to remind all students that they should not 100\% rely on the answers generated by AGI. And they should validate the generated answers with some external resources. Whether it's for building trust, enabling personalized learning, or ensuring ethical use, making AI interpretable is not just an added feature but a necessity in modern educational environments.

\subsubsection{Model Reproducibility and Replicability}

As AI systems become more complex and their applications more varied, the issues of Reproducibility and Replicability become increasingly important. \cite{goodchild2021replication} defined \textit{reproducibility} as the ability to obtain the same results using the same data and methods. Since there is no guarantee that different training procedures of the same neural network architecture on the same dataset will yield models which behave the same, there is potential for a big risk in applying these models in critical decision-making. A study by \cite{chao2020replicability} specifically addresses these concerns in the context of deep learning in software engineering, emphasizing the urgency of investigating reproducibility and replicability. The authors point out that only a small percentage of studies in the field even attempt to address these issues, which are critical for the credibility and generalizability of AI applications.

In the educational context, the importance of these factors cannot be overstated. For instance, a study by \cite{nelson2020notes} discusses the use of Google Colaboratory in AI education, highlighting its utility in teaching modern AI techniques interactively. The platform's ease of use and accessibility make it an ideal tool for educational purposes, but the reproducibility and replicability of the experiments conducted on it are crucial for both educators and students. Ensuring that experiments can be repeated by others not only validates the results but also makes the educational content more robust and reliable.

Moreover, the study by \cite{mulugeta2018credibility} on credibility, replicability, and reproducibility in simulation for biomedicine and clinical applications in neuroscience offers valuable insights that can be applied to AI in education. The authors emphasize the importance of best practices, verification, and validation to demonstrate model credibility to various stakeholders. In the educational setting, this would mean rigorous testing and validation of AI models used for teaching and assessment, ensuring that they meet the highest standards of credibility and ethical considerations.

In summary, the literature strongly advocates for a focus on the Reproducibility and Replicability of AI applications in education, particularly for AGI. Addressing these issues is not just a matter of academic integrity but is crucial for the practical application and widespread adoption of AGI in educational settings. Ensuring that AGI models are both reproducible and replicable will build trust among educators and learners, thereby facilitating more effective and ethical AGI-driven educational practices.

\subsubsection{Robot Rights}
The ethical implications of integrating robots into educational settings in the era of AGI are complex and multi-faceted, particularly regarding the question of robot rights. \cite{birhane2020robot} argue that the debate around robot rights often overshadows more pressing ethical concerns, such as machine bias and the exploitation of human labor. This suggests that while the concept of robot rights is intriguing, the focus should be on how the deployment of robots in educational settings impacts human welfare. Ethical considerations should prioritize the well-being of students and educators over the theoretical rights of machines. \cite{tavani2018can} reframe the question of robot rights to focus on whether some social robots qualify for moral consideration. This perspective is particularly relevant in educational settings where social robots are increasingly used for tutoring and emotional support. The ethical framework for these robots should consider their potential impact on students' moral and social development. Are these robots merely tools, or do they have a level of agency that requires ethical guidelines for their interaction with students?

In summary, while the debate on robot rights is intellectually stimulating, the immediate ethical concerns in educational settings revolve around human welfare, moral considerations, and the ethical education of students. As robots become more integrated into classrooms, addressing these issues through a well-rounded educational approach that includes ethical considerations is crucial. This will ensure that as technology advances, it is developed and used in a manner that is ethically sound and beneficial to human society.

\subsection{Codes of Conduct in the AGI Era}
While the ethical challenges surrounding using AGI in education are complex and evolving, they are not insurmountable. The literature provides a rich tapestry of insights into these issues, from data bias and model interpretability to the more philosophical questions surrounding robot rights. However, identifying the problems is only the first step; the next crucial phase involves developing actionable solutions that can be implemented in real-world educational settings. As we transition from understanding these ethical complexities to proposing solutions, it becomes evident that a multi-pronged approach is necessary. This involves the development of robust ethical frameworks and the practical steps for their implementation. In the following section, we outline a series of proposed solutions that address these ethical concerns comprehensively and promptly.

\begin{itemize}

\item \textbf{Ethical Auditing and Certification}. Given the complex ethical landscape surrounding the use of AGI in education, there is a pressing need for a standardized ethical auditing process. This would involve third-party organizations evaluating AGI systems against a set of ethical criteria, such as fairness, transparency, and data privacy. Completing the audit could result in an ethical certification, providing educators, students, and parents with assurance regarding the ethical integrity of the AGI system in use.

\item \textbf{Inclusive Stakeholder Consultation}. To ensure that the ethical frameworks governing AGI in education are comprehensive and equitable, involving a diverse range of stakeholders in their development is crucial. This could include educators, students, parents, ethicists, and representatives from marginalized communities. Their insights would help in tailoring ethical guidelines that are both practical and inclusive.

\item \textbf{Ethical AI Curriculum}. Building on the work of Williams (2021), educational institutions should integrate ethical AI education into their curricula. This would equip students with the critical thinking skills needed to navigate the ethical complexities of AGI, from data bias to robot rights. Such a curriculum could be developed with ethicists and AI researchers to ensure its comprehensiveness and relevance.

\item \textbf{Ethical Transition Plans}. As AGI technologies continue to advance, there is a growing concern about their potential to displace human roles in various sectors, including education. Ethical transition plans should be developed to manage this change responsibly. These could include retraining programs for educators and a phased integration of AGI technologies to allow for human oversight.

\item \textbf{Addressing Geographic Bias}. Data collection efforts should aim for global representation to mitigate geographic bias in AGI models. Collaborations could be formed with educational institutions worldwide to contribute to a more diverse and representative data set. This would help create more equitable AGI models and less likely to perpetuate existing social and geographic inequalities.

\end{itemize}

The ethical challenges posed by the integration of AGI in education are manifold but not insurmountable. Through a combination of ethical auditing, stakeholder consultation, educational initiatives, real-time monitoring, and transparency, it is possible to navigate these complexities responsibly. By taking a proactive and inclusive approach to ethical considerations, we can ensure that AGI serves as a force for good in educational settings, benefiting students, educators, and society.

\section{Discussion}\label{discussion}
As a game-changer, AGI has the potential to transform education by adapting to individual students' needs, creating personalized learning experiences, and improving assessment methods. The emergence of AGI has sparked a global debate on the future of education and the role of human educators \citep{zhai2023chatgpt1}. While AI technologies like ChatGPT and GPT-4 have the potential to revolutionize education, concerns have been raised about their impact on teacher roles \citep{edwards2018not}. Introducing LLMs also brings new challenges, such as resolving ethical issues, minimizing algorithmic biases, and ensuring responsible use in educational contexts \citep{sciencenewsChatGPTSimilar,latif2023knowledge}.

Despite its potential, AGI implementation in education faces challenges, such as the need for academic integrity and ensuring that the resources provided are culturally relevant and appropriate. However, by addressing these concerns, AGI can revolutionize the educational landscape, creating more inclusive, personalized, and effective learning experiences for all students.

To address ethical concerns, researchers, educators, and learners must collaborate as AGI develops to maximize its potential for improving education while addressing the accompanying risks and uncertainties. Eight principles have been proposed \citep{ethicalInstituteEthical}, including human augmentation, bias evaluation, explainability, reproducible operations, displacement strategy, practical accuracy, trust by privacy, and data risk awareness. These principles help ensure the responsible development of AGI and can be used to guide the implementation of AI in education. In addition, codes of conduct for the use of AGI in academia should be developed to address ethical issues, and regularity in the use of AGI in homework, teaching, and recruitment should be established to ensure fairness and prevent cheating.

Data bias and fairness are also significant ethical concerns in AGI development, particularly in educational settings. AI models can produce incorrect results if the training data used is unbalanced or incomplete. To mitigate this risk, it is crucial to ensure that the data used to train AGI models is diverse and representative of all students, regardless of race, gender, socioeconomic status, or other demographic factors. Educators must ensure that AGI-driven systems do not perpetuate biases or discrimination in their assessments or recommendations. Additionally, \cite{zhai2022pseudo} warned that pseudo-AI bias could harm the public's acceptance of AI.

Privacy is another critical consideration in AGI development. For example, AGI systems may access sensitive health information without the user's consent, and it is crucial to establish policies and procedures to protect student privacy in educational settings \citep{nabi2018bioethics}. This includes ensuring that student data is only used for educational purposes and is not shared or sold to third parties without explicit consent.

Model interpretability and reproducibility are essential to evaluate the reliability of AGI results. Developers must ensure that AGI models are transparent and explainable, enabling educators to understand how the system arrives at its assessments and recommendations. Reproducibility is also crucial, allowing educators to replicate and verify the results of AGI systems \citep{goodchild2021replication}.
The displacement strategy is a growing concern as AGI may replace humans in various occupations, including education. While AGI has the potential to revolutionize education, it is essential to design a curriculum to prepare students for these changes. This includes teaching students how to use and interact with AGI systems effectively and preparing them for the changing job market. Additionally, educators must ensure that AGI-driven systems do not replace human interaction and expertise in education.

Concerns have also been raised about the impact of AGI on teacher roles. While some fear AI might replace teachers, it is essential to recognize the value of human interaction and expertise in education \citep{zhai2022pseudo}. In addition, teachers provide guidance, motivation, and emotional support that AI cannot fully replicate. Therefore, AI should be viewed as a tool to complement and enhance teachers' capabilities rather than threaten their profession.

Robot rights are another critical consideration of AGI ethics, as AGI may raise questions about whether robots should possess the same rights as humans \citep{wikipediaEthicsArtificial}. While this issue is still under debate, it is crucial to ensure that AGI systems are developed and used in a manner that is consistent with ethical and moral principles.

Codes of conduct for using AGI in academia should be developed to address ethical issues. Regular use of AGI in homework, teaching, and recruitment should be established to ensure fairness and prevent cheating \citep{icmlICML2023}. Educators and students must be aware of the potential ethical implications of AGI use and ensure that AGI systems are developed and used in a manner that is consistent with ethical principles.

\section{Conclusion}
\label{conclusion}
The paper presented evidence for the potential of AGI to revolutionize education by providing personalized instruction, adaptive learning pathways, and instant feedback, contributing to more effective learning outcomes. AGI can also enhance teachers' pedagogical content knowledge, allowing for differentiated instruction and support for diverse student needs. The paper also highlighted that By utilizing potential AGI-like chatbots, teachers can access a wealth of information and activities catering to various learning styles, leading to more targeted lesson plans and teaching strategies. Additionally, AGI algorithms can support Human-Computer Interaction (HCI), transforming education by enhancing the design and usability of digital tools, improving the accessibility of learning resources, and creating engaging and interactive learning experiences for students. Finally, the paper mentioned new challenges in implementing AGI for education, such as resolving ethical issues, minimizing algorithmic biases, and ensuring responsible use in educational contexts. The paper suggests that educators, researchers, and learners must collaborate as AGI develops to maximize its potential for improving education while addressing the accompanying risks and uncertainties.


\section*{Declaration of generative AI and AI-assisted technologies in the writing process}

During the preparation of this work the authors used ChatGPT in order to check grammar and polish the wordings. After using this tool/service, the authors reviewed and edited the content as needed and take full responsibility for the content of the publication.

\bibliographystyle{cas-model2-names}

\bibliography{cas-refs}



\end{document}